\documentclass[final,3p,times,twocolumn]{elsarticle}

\usepackage{amsmath}
\usepackage{amssymb}
\usepackage{amsthm}
\usepackage{booktabs}
\usepackage{cancel}
\usepackage{caption}
\usepackage{subcaption}
\usepackage{chemarrow}
\usepackage{chemformula}
\usepackage{chngcntr}
\usepackage{color}
\usepackage{colortbl}
\usepackage{cuted}
\usepackage{dblfloatfix}
\usepackage{enumitem}
\usepackage{epsfig}
\usepackage{epstopdf}
\usepackage{esint}
\usepackage{etoolbox}
\usepackage{geometry}
\usepackage{graphics}
\usepackage{graphicx}
\usepackage{harpoon}
\usepackage{hyperref}
\usepackage{latexsym}
\usepackage{lineno}
\usepackage{lipsum}
\usepackage{longtable}
\usepackage{lscape}
\usepackage{mathrsfs}
\usepackage{mathtools}
\usepackage[version=4]{mhchem}
\usepackage{multicol}

\usepackage{soul}
\usepackage{threeparttable}
\usepackage{tikz}
\usepackage[tight,nice]{units}
\usepackage{xcolor}

\DeclareMathOperator*{\argminA}{arg\,min}

\setlist[itemize]{noitemsep, nolistsep}
\definecolor{LightCyan}{rgb}{0.88,1,1}
\definecolor{Gray}{gray}{0.85}
\hypersetup{
    colorlinks=false,
    linkcolor=black,
    filecolor=magenta,      
    urlcolor=black,
    }

\biboptions{sort&compress}
\makeatletter
\def\ps@pprintTitle{%
  \let\@oddhead\@empty
  \let\@evenhead\@empty
  \let\@oddfoot\@empty
  \let\@evenfoot\@oddfoot
}

\newcommand{\revone}[1]{\textcolor{black}{{#1}}}
\newcommand{\revtwo}[1]{ \textcolor{black}{{#1}}}
\newcommand{\revall}[1]{ \textcolor{black}{{#1}}}
\newcommand{\revtwotwo}[1]{ \textcolor{black}{{#1}}}
\newcommand{\revtwothree}[1]{ \textcolor{black}{{#1}}}

\newrobustcmd*{\myVtriangle}[2]{\tikz{\filldraw[draw=#1,fill=#2] (0cm,0.2cm) --
(0.2cm,0.2cm) -- (0.1cm,0cm) -- (0cm,0.2cm);}}

\newrobustcmd*{\mythickVtriangle}[2]{\tikz{\filldraw[line width=0.3mm,draw=#1,fill=#2] (0cm,0.2cm) --
(0.2cm,0.2cm) -- (0.1cm,0cm) -- (0cm,0.2cm);}}

\newrobustcmd*{\mythickErrorVtriangle}[2]{\tikz{\filldraw[line width=0.3mm,draw=#1,fill=#2] (-0.05cm,0.05cm) --
(0.05cm,0.05cm) -- (0cm,-0.05cm) -- (-0.05cm,0.05cm);  \draw[draw=#1] (0.0cm, -0.12cm) -- (0.0cm, 0.12cm) ; \draw[draw=#1] (-0.06cm, 0.12cm) -- (0.06cm, 0.12cm); \draw[draw=#1] (-0.06cm, -0.12cm) -- (0.06cm, -0.12cm)    }}

\newrobustcmd*{\mytriangle}[2]{\tikz{\filldraw[draw=#1,fill=#2] (0.0cm,0.0cm) --
(0.2cm,0cm) -- (0.1cm,0.2cm) -- (0cm,0cm);}}

\newrobustcmd*{\mysquare}[2]{\tikz{\draw[draw=#1,fill=#2] (0cm,0cm)
rectangle (0.2cm,0.2cm)}}

\newrobustcmd*{\mythicktriangle}[2]{\tikz{\filldraw[line width=0.3mm,draw=#1,fill=#2] (0.0cm,0cm) --
(0.2cm,0cm) -- (0.1cm,0.2cm) -- (0.0cm,0cm);}}

\newrobustcmd*{\mythicksquare}[2]{\tikz{\draw[line width=0.3mm,draw=#1,fill=#2] (0cm,0cm)
rectangle (0.2cm,0.2cm)}}

\newrobustcmd*{\mybarredtriangle}[2]{\tikz{\draw[draw=#1,fill=#2] (0,0) --
(0.2cm,0) -- (0.1cm,0.2cm) -- (0cm,0cm); \draw[draw=#1] (-0.1cm, 0.07cm) -- (0.3cm, 0.07cm)}}

\newrobustcmd*{\mythickbarredtriangle}[2]{\tikz{\draw[line width=0.3mm,draw=#1,fill=#2] (0,0) --
(0.2cm,0) -- (0.1cm,0.2cm) -- (0cm,0cm); \draw[draw=#1] (-0.1cm, 0.07cm) -- (0.3cm, 0.07cm)}}

\newrobustcmd*{\mybarredsquare}[2]{\tikz{\draw[draw=#1,fill=#2] (0,0)
rectangle (0.2cm,0.2cm); \draw[draw=#1] (-0.1cm, 0.1cm) -- (0.3cm, 0.1cm)}}

\newrobustcmd*{\mythickbarredsquare}[2]{\tikz{\draw[line width=0.3mm,draw=#1,fill=#2] (0,0)
rectangle (0.2cm,0.2cm); \draw[draw=#1] (-0.1cm, 0.1cm) -- (0.3cm, 0.1cm)}}

\newrobustcmd*{\mybarredcircle}[2]{\tikz{\draw[draw=#1,fill=#2] (0,0)
circle (0.1cm); \draw[draw=#1] (-0.2cm, 0.0cm) -- (0.2cm, 0.0cm)}}

\newrobustcmd*{\mythickbarredcircle}[2]{\tikz{\draw[line width=0.3mm,draw=#1,fill=#2] (0,0)
circle (0.1cm); \draw[draw=#1] (-0.2cm, 0.0cm) -- (0.2cm, 0.0cm)}}

\newrobustcmd*{\mythickErrorcircle}[2]{\tikz{\draw[line width=0.3mm,draw=#1,fill=#2] (0,0)
circle (0.06cm); \draw[draw=#1] (0.0cm, -0.12cm) -- (0.0cm, 0.12cm) ;   \draw[draw=#1] (-0.06cm, 0.12cm) -- (0.06cm, 0.12cm); \draw[draw=#1] (-0.06cm, -0.12cm) -- (0.06cm, -0.12cm)    }}

\newrobustcmd*{\mydashedline}[1]{\tikz{\draw[draw=#1] (-0.2cm, 0.2cm) -- (-0.1cm, 0.2cm); \draw[draw=#1] (-0.0cm, 0.2cm) -- (0.1cm, 0.2cm)}}

\newrobustcmd*{\mythickcross}[1]{\tikz{\draw[line width=0.3mm,draw=#1] (0,0) --
(0.2cm,0); \draw[line width=0.3mm,draw=#1] (0.1cm,-0.1cm) -- (0.1cm,0.1cm);}}

\newrobustcmd*{\mybarredcross}[1]{\tikz{\draw[line width=0.3mm,draw=#1] (0,0) --
(0.2cm,0); \draw[line width=0.3mm,draw=#1] (0.1cm,-0.1cm) -- (0.1cm,0.1cm); \draw[draw=#1] (-0.1cm,0) -- (0.3cm,0);}}

\newrobustcmd*{\myline}[1]{\tikz{\draw[draw=#1] (-0.15cm, 0.1cm) -- (0.15cm, 0.1cm);\draw[line width=0.3mm,draw=#1] (-0.0cm, 0.0cm);}}

\newrobustcmd*{\mythickline}[1]{\tikz{\draw[line width=0.3mm,draw=#1] (-0.15cm, 0.1cm) -- (0.15cm, 0.1cm);\draw[line width=0.3mm,draw=#1] (-0.0cm, 0.0cm);}}

\newrobustcmd*{\mythickdashedline}[1]{\tikz{\draw[line width=0.3mm,draw=#1] (-0.2, 0.1cm) -- (-0.1cm, 0.1cm); \draw[line width=0.3mm,draw=#1] (-0.0cm, 0.1cm) -- (0.1cm, 0.1cm); \draw[line width=0.3mm,draw=#1] (-0.0cm, 0.0cm);}}

\newrobustcmd*{\mythickdasheddottedline}[1]{\tikz{\draw[line width=0.3mm,draw=#1] (-0.22, 0.1cm) -- (-0.13cm, 0.1cm); \draw[line width=0.3mm,draw=#1] (-0.085cm, 0.1cm) -- (-0.055cm, 0.1cm); \draw[line width=0.3mm,draw=#1] (-0.01cm, 0.1cm) -- (0.08cm, 0.1cm); \draw[line width=0.3mm,draw=#1] (-0.0cm, 0.0cm);}}

\newrobustcmd*{\mycircle}[2]{\tikz{\draw[draw=#1,fill=#2] (0,0)
circle (0.1cm);}}

\newrobustcmd*{\mythickcircle}[2]{\tikz{\draw[line width=0.3mm,draw=#1,fill=#2] (0,0)
circle (0.1cm);}}

\newrobustcmd*{\mydot}[1]{\tikz{\draw[line width=0.3mm,draw=#1] (0,0)
circle (0.025cm);}}

\journal{J. of Energy Storage}

\begin{document}

\begin{frontmatter}

\title{PINN surrogate of Li-ion battery models for parameter inference.\\Part I: Implementation and multi-fidelity hierarchies for the single-particle model}

\author[NREL_CompSci]{Malik Hassanaly\corref{cor}}
	\cortext[cor]{Corresponding author. Tel: (303) 275-4739.}
		\ead{malik.hassanaly@nrel.gov}
\author[NREL_ECaSS]{Peter J. Weddle}
\author[NREL_CompSci]{Ryan N. King}
\author[NAU]{Subhayan De}
\author[UC_Boulder]{Alireza Doostan}
\author[NREL_ECaSS]{\\Corey R. Randall}
\author[INL]{Eric J. Dufek}
\author[NREL_ECaSS]{Andrew M. Colclasure}
\author[NREL_ECaSS]{Kandler Smith}

\address[NREL_CompSci]{Computational Science Center, National Renewable Energy Laboratory (NREL), Golden, CO 80401}
\address[NREL_ECaSS]{Energy Conversion and Storage Systems Center, National Renewable Energy Laboratory, Golden, CO 80401}
\address[UC_Boulder]{Aerospace Mechanics Research Center, University of Colorado, Boulder, CO 80303}
\address[NAU]{Mechanical Engineering Department, Northern Arizona University, Flagstaff, AZ 86011}
\address[INL]{Energy Storage and Electric Transportation Department, Idaho National Laboratory (INL), Idaho Falls, ID 83415}


\begin{abstract} 
To plan and optimize energy storage demands that account for Li-ion battery aging dynamics, techniques need to be developed to diagnose battery internal states accurately and rapidly.  This study seeks to reduce the computational resources needed to determine a battery's internal states by replacing physics-based Li-ion battery models -- such as the single-particle model (SPM) and the pseudo-2D (P2D) model -- with a physics-informed neural network (PINN) surrogate.  The surrogate model makes high-throughput techniques, such as Bayesian calibration, tractable to determine battery internal parameters from voltage responses. This manuscript is the first of a two-part series that introduces PINN surrogates of Li-ion battery models for parameter inference (i.e., state-of-health diagnostics).  In this first part, a method is presented for constructing a PINN surrogate of the SPM.  A multi-fidelity hierarchical training, where several neural nets are trained with multiple physics-loss fidelities is shown to significantly improve the surrogate accuracy when only training on the governing equation residuals.  The implementation is made available in a companion repository  (\hyperlink{https://github.com/NREL/PINNSTRIPES}{https://github.com/NREL/PINNSTRIPES}).  The techniques used to develop a PINN surrogate of the SPM are extended in Part II \revone{\cite{hassanaly2023pinn2}} for the PINN surrogate for the P2D battery model, and explore the Bayesian calibration capabilities of both surrogates.

\end{abstract}


\end{frontmatter}

\section{Introduction}
\label{sect:Introduction}

Electrochemical storage technology is an important part of the decarbonization transition. Integrating electrochemical storage solutions into the power grid is expected to improve strategic deployment, stabilize the grid, and peak-shift energy supply/demand~\cite{TZRSKWKSOCLRPC21}. Additionally, electrochemical storage solutions are used in vehicles (e.g., \ce{Li}-ion batteries) to enable cleaner alternatives to fossil fuels. \ce{Li}-ion batteries are a particularly successful electrochemical storage device in grid and consumer applications, including electric vehicles~\cite{RMBD22,TWYCCFLPKAUCBDSCWWY22,DABCCCGKKMRRSTUW22,HHTCTJBP21,HSKJ17}. A key challenge in \ce{Li}-ion battery technology is retaining high-energy density and high-power capability, while simultaneously improving cycle-life and calendar-life. However, \ce{Li}-ion battery lifetime and aging dynamics vary significantly with chemistry, operating conditions, cycling demands, electrode design, and operational history, which makes optimal handling, design, and maintenance difficult~\cite{WKCYGCSGTD23,ABPDDDGGGHLKPSSSS22,GGDS21}.  

To determine the optimal \ce{Li}-ion battery usage for maximizing the lifetime, rapid assessment and forecasting of battery state-of-health are required.  In the battery research community, many researchers use machine learning approaches to assess remaining life and diagnose battery age-states~\cite{GGDS21,WKCYGCSGTD23,SGCSY21,GSSSY22,KYCTD22,CSAD22}.  However, these machine-learning approaches typically rely on testing a significant number of cells to obtain enough data to accurately project cycling/calendaring fade~\cite{GGDS21,GSSSY22}, or require slow (on the order of 20--40~h) reference performance tests (RPTs) to diagnose a battery's aged state~\cite{SGCSY21,CSAD22,KYCTD22}.  In contrast, our approach reduces the data requirements by additionally leveraging \revtwothree{physics-based constraints \cite{aykol2021perspective,tu2023integrating} that use} the community-accepted governing equations that describe the internal kinetic/transport physics.  With a machine-learning model that approximates the solutions of the governing equations, internal battery parameters can be extracted by analyzing high-rate (2~C) voltage responses. Typically, the internal batter parameters include transport parameters, the initial battery state and parameters that characterize the reaction kinetics. In the present study, the battery's internal states are determined by using well-accepted surrogates of \ce{Li}-ion physics-based models, including the single-particle model (SPM) (in Part~I) and the pseudo-2D (P2D) model~\cite{FDN94,SGW07} (in Part~II \revone{\cite{hassanaly2023pinn2}}). The physics information that complements the lack of data is included by constructing a physics-informed neural network (PINN) \cite{raissi2019physics}.

The primary purpose of developing PINNs for \ce{Li}-ion batteries is to drastically decrease the computational time required to solve these physics-based models.  Once a PINN surrogate is trained, it can solve the SPM in the order of $10^5$ faster and the P2D model in the order of $10^6$ faster, as compared to using a partial differential equation (PDE) solver. Notoriously, PINN training can be prone to instabilities \cite{krishnapriyan2021characterizing,wang2021understanding}, which are addressed in this work. Training a PINN is computationally more expensive than solving a set of PDEs once. However, when using the PINN with techniques such as Bayesian calibration and Markov-Chain Monte-Carlo (MCMC), multiple PDE solutions need to be generated, which compensates for the initial training cost. These techniques are especially useful to inversely determine the battery's internal state from voltage trajectories and help diagnose the battery's state-of-health (and confidence intervals of these states) using high-rate voltage responses.

%
\subsection{Previous Li-ion degradation modeling}
Modeling \ce{Li}-ion battery degradation can be done at a macroscopic level where the battery lifetime is simulated as a function of high-level parameters such as operating conditions (e.g., C-rate, depth-of-discharge, temperature, equivalent full cycles) and electrode composition~\cite{PKWSLB22,GGDS21,SGCSY21,GSSSY22,GCHJS22}. However, such methods give, by construction, macroscopic, cell/build-specific information about battery degradation, which is valuable but can prevent transferability of the model to new battery types, and hinder the physical interpretability of the internal degradation mechanisms~\cite{GCHJS22,PKWSLB22}. Instead, it can be advantageous to introduce more granularity in the modeling framework by capturing internal property dynamics during aging~\cite{BFSV23,LDCJRJS22,GZZG21,ASKKKEJL22,WKCYGCSGTD23,MSUKUYU23,EKPKPHGACYLPDTMRWPWO21,OAMATSEWOM22}. This approach requires estimating multiple relevant battery internal parameters during cycling. Despite recent advances in experimental battery diagnostics, many internal parameters such as effective solid-phase diffusivity, electrolyte transport properties, surface kinetic rates, and particle surface/diffusion length can be difficult to measure/discern without destructive test, which would prevent further cycling of a given cell~\cite{FSDKJC22}. Non-destructive tests (e.g., RPTs, and electrochemical impedance spectroscopy) can also affect, even mildly, some of the battery dynamics, such as cycle-by-cycle polarization, thereby introducing noise within the analysis~\cite{CKTD21}.  Thus, there exists a need to develop tools that can rapidly determine a battery's internal state in a non-destructive way and ideally without changing the battery's cyclic demands.

\subsection{PINN surrogate model}

With the primary goal of developing a surrogate \ce{Li}-ion battery model for enabling fast calibration of the battery's internal parameters\revtwothree{,} there are at least three features that need to be met: 
\begin{itemize}
    \item[1)] The surrogate model needs to replicate the physics-based model (e.g., the SPM and the P2D model). 
    \item[2)] The surrogate model needs to be accurate over the entire space where it will be interrogated, i.e., the spatiotemporal space where observational data is gathered (of dimension 2 or 3, depending on the physics model), and the parametric space being explored as part of the Bayesian calibration (of dimension possibly greater than 20 \cite{reddy2019accelerating}). 
    \item[3)] The surrogate model must be significantly more computationally efficient as compared to the physics-based model.
\end{itemize}
In the present study, a surrogate PINN is used to approximate the physics-based models, while still capturing the observable response of internal battery parameters of interest.  

When selecting a particular data-driven model, a primary consideration is the amount/quality of available data available for training.  In the case of physics-based \ce{Li}-ion battery models, it is reasonable to expect that a large number of data points that span the spatiotemporal domain can be obtained via traditional PDE solvers (i.e., use solutions of the physics-based models to train a data-driven model). However, it is unreasonable to expect that a sufficient number of PDE solutions can be generated to span the full parametric domain.  Ideally, the surrogate model would be designed to handle a large amount of data, and be accurate even in the absence of data. These needs lead to developing a PINN surrogate model to approximate physics captured in typical \ce{Li}-ion battery models.

In the rest of the work, the term PINN refers to using the governing equation residuals to train a neural network similar to the approach in Raissi et al.~\cite{raissi2019physics}, rather than using only training data originating from physics-based simulations~\cite{li2021physics}. In previous works, PINNs capturing \ce{Li}-ion battery physics or redox-flow battery physics~\cite{CHEN2023233548} were used to enforce physics constraints to obtain inferred parameters as the neural network output~\cite{nascimento2021hybrid, singh2023hybrid,he2022physics}. This approach does not require the PINNs to be accurate over the entire parametric domain. However, they must be retrained each time the calibration data set changes. \revtwothree{Zheng et al.~\cite{zheng2023inferring} addressed this issue with a similar strategy as the one used here~\cite{hassanaly2022physics}.  That is, by feeding the parameters identified as the PINN inputs. Compared to this work, Zheng et al.~\cite{zheng2023inferring} does not discuss the effect of training choices involved with physics-informed losses, which is the main focus of Part~I.}

In the present approach, the PINN is trained once and can be reused with any new data set. Unlike traditional supervised neural nets, PINNs can rely on the governing equations of the system themselves to complement the lack of data. However, standard PINNs are notoriously expensive to train \cite{grossmann2023can} and are subject to multiple instabilities \cite{krishnapriyan2021characterizing,wang2021understanding}. The main contributions of \revtwothree{the present manuscript} are as follows:
\begin{itemize}
    \item We find that residual blocks and merged neural-net architectures promote high accuracy of the PINN in a low-data regime. 
    \item \revtwothree{We evaluate and discuss the efficacy of several PINN training regularization procedures}.
    \item \revtwothree{We evaluate the effect of linearizing the Butler--Volmer kinetics on the PINN accuracy.}
    \item \revtwothree{We derive a multi-fidelity training procedure that improves the PINN accuracy. We demonstrate its benefit when using non-linear Butler--Volmer kinetics}.
\end{itemize}

The \revtwothree{first and fourth bullets} related to the architecture and multi-fidelity training are discussed in the present manuscript in relation to the SPM (Part~I).  These attributes are further used in Part~II \revone{\cite{hassanaly2023pinn2}} to develop PINNs that can capture the P2D model physics. \revtwothree{Applying PINNs for parameter identification is also discussed in Part II.}


\subsection{Manuscript organization}

The present work is divided into two parts.  In Part~I, the PINN training procedure, weight initialization, architecture effects, and regularization techniques are explored using SPM governing equations. A multi-fidelity training procedure is shown to address training instabilities observed when using the SPM equation residuals. In Part~II \revone{\cite{hassanaly2023pinn2}}, the PINN is extended to solving the P2D model equations.  In general, minimizing the P2D governing equation residuals is significantly more difficult as compared to training a PINN to solve the SPM.  These difficulties were addressed by using a novel training loss regularization which reflects domain-specific knowledge about battery electrochemistry.


\section{Single-particle model}
\label{sec:phyModels}

The single-particle model is a standard model in the \ce{Li}-ion battery community~\cite{SGRW06}.  The model captures solid-phase \ce{Li} transport resistances and electrochemical overpotentials at the electrolyte/electrode interfaces~\cite{SGRW06,DWZCBJK18,GSW11,MAKMK17}.  The model assumes: 1) the reactive electrolyte/electrode surface area is well approximated by a collection of disconnected spheres, 2) the composite electrode solid-phase transport is well approximated by Fickian diffusion within a single, spherical particle, 3) the electrolyte is ``ideal'' where ionic concentration $c_{\rm e}$ is constant and the potential $\phi_{\rm e}$ is uniform, and 4) the electrode potential $\phi_{{\rm s},j}$ within each composite electrode is assumed to be uniform (i.e., $\phi_{{\rm s},j} (t)$).  The model is most appropriate for studying low-rate battery responses, where these simplifying assumptions are reasonable~\cite{SGRW06}.

In the single-particle model, there are two independent variables: the \ce{Li} concentration in the anode particle $c_{\rm s, an}(r,t)$ and the \ce{Li} concentration in the cathode particle $c_{\rm s, ca}(r,t)$.  These concentrations are assumed to follow Fick's law as
\begin{equation}
    \frac{\partial c_{{\rm s},j}}{\partial t} = \frac{1}{r^2}\frac{\partial }{\partial r}\left(D_{{\rm s},j} r^2\frac{\partial c_{{\rm s},j}}{\partial r}\right),
\end{equation}
where $j$ indicates either the anode or cathode domain, $D_{\rm{s}}$ is the \ce{Li} solid-phase diffusivity, $r$ is the partial radial direction, and $t$ is time.  To initiate the simulation, the solid-phase concentration for either electrode is assumed to be spatially uniform as $c_{{\rm s},j}(r)|_{t=0} = c_{{\rm s},0,j}$, where $c_{{\rm s},0,j}$ is the initial concentration. At the particle center, the flux is zero due to symmetry.  At the particle surface, the flux due to reactions is
\begin{equation}
    \label{eq:fluxion}
    \left(D_{{\rm s},j} \frac{\partial c_{{\rm s},j}}{\partial r}\right)_{r=R_j} = - J_j,
\end{equation}
where $R_j$ is the particle radius and $J_j$ is the flux of ions from the surface.  The flux of ions at the surface is dictated by the current demand $I$, which can be expressed as
\begin{equation}
    \label{eq:current_and_flux}
    J_{\rm ca} = \frac{I}{F}\frac{R_{\rm ca}}{3\epsilon_{\rm ca}V_{\rm ca}}, \quad J_{\rm an} = \frac{-I}{F}\frac{R_{\rm an}}{3\epsilon_{\rm an}V_{\rm an}},
\end{equation}
where $F$ is Faraday's constant, $\epsilon_j$ is the active material volume fraction, and $V_j$ is the total composite electrode volume.  It is common to extract the voltage response from this set of decoupled ordinary differential equations (ODEs) by assuming the intercalation reactions follow the Butler--Volmer expression on either electrode
\begin{equation}
\label{eq:BV}
\begin{split}
    J_j = \frac{i_{0,j}}{F}\Bigg[&\exp\left(\frac{\alpha_{\rm a}F\left(\phi_{{\rm s},j}-\phi_{\rm e}-U_{{\rm OCP},j}\right)}{RT}\right) \\&-\exp\left(\frac{(\alpha_{\rm a}-1) F\left(\phi_{{\rm s},j}-\phi_{\rm e}-U_{{\rm OCP},j}\right)}{RT}\right)\Bigg],
\end{split}
\end{equation}
where the exchange current density $i_{0,j}$ can be expressed as
\begin{equation}
    i_{0,j} = i^0_{0,j} c_{\rm e}^{\alpha_{\rm a}} \left(c_{{\rm s,max},j} - c_{{\rm s},j}|_{r=R_j}\right)^{\alpha_{\rm a}} \left(c_{{\rm s},j}|_{r=R_j}\right)^{(1-\alpha_{\rm a})}.
\end{equation}
Here, $i^0_{0,j}$ is the exchange current density prefactor, $c_{\rm e}$ is the time-independent, uniform \ce{Li}-ion concentration in the electrolyte, and $c_{{\rm s,max},j}$ is the maximum electrode concentration.  The anodic transfer coefficient $\alpha_{\rm a}$ is typically assumed to be 0.5 in \ce{Li}-ion battery models.  Although the open-circuit voltage of the electrode is related to the reactant/product thermodynamics~\cite{CK10}, in practice, the open-circuit voltage of each electrode $U_{{\rm OCP},j} (c_{{\rm s},j})$ is a measured, tabulated value that depends on the surface solid-phase concentration $c_{{\rm s},j}|_{r=R_j}$.  Finally, by setting an electrode potential to reference (i.e., $\phi_{\rm s,an}$ = 0) and doing some algebraic manipulations, the battery voltage can be determined from $\phi_{\rm s,ca}-\phi_{\rm s,an}$.  The single-particle formulation is presented sparingly here as this is a standard model in the \ce{Li}-ion battery community.  Detailed derivations of the single-particle model are provided elsewhere~\cite{DWZCBJK18,GSW11,SGRW06,MAKMK17}.

In the present work, a PINN is developed as a surrogate model for the discharge of a single-particle model at 2~C. The discharge is modeled over the time-interval of $[0, 1350~{\rm s}]$ to avoid the need to model the large temporal gradient of the positive electrode potential near the very end of discharge. Modeling the discharge (rather than the charge) requires enforcing only a constant-current discharge which is easier to enforce than a constant-current, constant-voltage charge, where the boundary conditions need to be varied over time. The SPM model parameters are chosen from a well-studied cell~\cite{CTJTDPBRFEDS20}. The role of parameter uncertainty in the response of an SPM model has been studied elsewhere; see, e.g.,~\cite{HMD15,CD17}. It should be noted that the SPM is considered a computationally inexpensive model in the energy storage community~\cite{SGRW06}.  However, there is value in developing PINNs to further increase the computational efficiency of the SPM model in the context of parameter estimation and sensitivity. As will be shown in Part~II \revone{\cite{hassanaly2023pinn2}}, an SPM surrogate is also useful to speed up the training of PINN surrogates of higher fidelity \ce{Li}-ion battery models (i.e., the P2D model). 

\section{Methods: PINN for battery models}
\label{sec:models}

In this work, an artificial neural network is used to approximate the mapping from spatiotemporal variables (here $t$ for the temporal variable and $r$ for the spatial variable) to the battery state variables $\xi(t,r)$. In the case of the SPM, $\xi$ denotes either the concentration of \ce{Li} in the anode $c_{\rm s,an}$, the concentration of \ce{Li} in the cathode $c_{\rm s,ca}$, the potential in the electrolyte $\phi_{\rm e}$,  or the potential at the cathode current collector $\phi_{\rm s,ca}$. Artificial neural networks (referred to as neural networks in the rest of the manuscript) mimic biological neural networks in that they use layers of neurons activated by non-linear functions. The appeal of neural networks is partly due to the fact that they are differentiable with respect to any of their parameters, and are provably able to approximate any functional form~\cite{hornik1989multilayer}. Traditionally, neural networks are trained with a data-based approach, i.e., by showing the neural network sufficiently many pairs $\{(t,r), \xi\}$, a sufficiently large neural network can approximate an underlying function $\xi(t,r)$. In the case where the function $\xi(t,r)$ can be approximated by solving governing equations, as is the case of \ce{Li}-ion battery models, the neural network need not rely solely on input/output data pairs but can also learn from the governing equations themselves. 

Physics-based governing equations typically take the form 
\begin{equation}
    \label{eq:gov}
    \mathcal{R}\left(\xi(t,r)\right) = 0,
\end{equation}
where $\mathcal{R}$ denotes the governing equations' residual. Appropriate initial and boundary conditions are also assumed. In general $\mathcal{R}$ involves derivatives with respect to the spatiotemporal variables. If the spatiotemporal variables are used as input or parameters of the neural networks, the residual $\mathcal{R}\left(\xi(t,r)\right)$ can be readily evaluated at any spatiotemporal location $(t,r)$, thanks to the auto-differentiation capability of the neural network. In turn, Eq.~\ref{eq:gov} can be used as a constraint to train the neural network, in place of data~\cite{sun2020surrogate}, or to supplement a data-based approach~\cite{raissi2019physics}. This method is known as physics-informed neural networks (PINNs) and refers to using a physics-informed loss function. The advantage of this approach, of particular interest here, is that PINNs can use a data-based approach where data is available and compensate for the lack of data with governing equation residuals where data is not available or is scarce. \revone{If all the governing equations and boundary conditions are known, PINNs can even be trained without any data \cite{koric2023data,harandi2024mixed,lu2021learning}.}

PINNs use a mixture of data (input and output pairs available) $\mathcal{D}$ and residual evaluations at collocation points $\mathcal{C}$ during training. The collocation points are placed as needed throughout the spatiotemporal and parametric domains, where one would like to enforce that the governing equations need to be satisfied. At the collocation points, the residual (Eq.~\ref{eq:gov}) of the governing equations (cf.~Section~\ref{sec:phyModels}) is minimized. At the data points, the mismatch between the input/output pair predicted and shown to the network is minimized~\cite{raissi2019physics}.

In line with typical approaches \cite{raissi2019physics}, the PINN approximates the state variables $\xi(t,r)$ with a neural network parameterized by a set of weights and biases $\boldsymbol{\theta}$, by minimizing a global loss function $\mathcal{L}$.  Formally, the optimization problem can be written as 

\begin{equation}
    \label{eq:opt}
    \argminA_{\boldsymbol{\theta}} \mathcal{L}(\mathcal{C}, \mathcal{D}, \boldsymbol{\theta})
\end{equation}
where the global loss $\mathcal{L}$ is calculated using the output of PINN which depends $\boldsymbol{\theta}$. The global loss can be decomposed as 
\begin{equation}
    \label{eq:totalLoss}
    \mathcal{L} = \mathcal{L}_{\rm int} (\mathcal{C}, \boldsymbol{\theta}) + \mathcal{L}_{\rm bound} (\mathcal{C}, \boldsymbol{\theta}) + \mathcal{L}_{\rm data} (\mathcal{D}, \boldsymbol{\theta}), 
\end{equation}
where $\mathcal{L}_{\rm int}$ is the average of the mean squares error of the residual at the collocation points located in the interior of the domain, $\mathcal{L}_{\rm bound}$ is the mean squares error of the residual at the collocation points located at the spatial boundaries of the domain, and $\mathcal{L}_{\rm data}$ is the mean squared error of the predicted state variables values against available data. \footnote{The sum of the first two terms relating to errors from the governing equations is commonly referred to as ``physics loss''.  If there is no data, the ``global loss'' and the ``physics loss'' are equivalent.}\revone{The residuals contribution to the loss functions are also weighted in the global loss function. An extensive discussion about the choice of the residual weights is provided in Sec.~\ref{sec:BalancingPhysicsTermsWithHyperparameters}.}

\subsection{Strict enforcement of initial \& boundary conditions}
\label{sec:hardEnf}
Following Ref.~\cite{sun2020surrogate,sukumar2022exact}, the initial conditions are not enforced via a loss term, but via distance functions, which guarantees that the PINN exactly matches the prescribed initial conditions. The physical value of any state variable $\xi(t)$ is predicted as 
\begin{equation}
    \label{eq:dist}
    \xi(t) = \widetilde{\xi}(t) F(t) + \xi_{0},
\end{equation}
\revtwo{where $\widetilde{\xi}(t)$ is the raw output of the neural net, $F(t) = 1 - \exp(-t/\tau)$}, $\tau$ is a timescale over which the initial condition has a significant effect, and $\xi_0$ is the initial condition to enforce. In all the situations discussed below, $\tau$ is set to $1$~s. This approach ensures that at $t=0$~s the predicted solution exactly matches the prescribed initial condition (i.e., $\xi(0) = \xi_{0}$). At later times, the predicted solution relaxes to the output of the neural net $\widetilde{\xi}(t)$. This approach is applicable to all state variables in the SPM since their initial values are specified for the parabolic, ordinary differential equations. In the P2D model, all the variables can be treated similarly aside from the potentials (i.e., the algebraic constraint equation) which is discussed extensively in Part~II \revone{\cite{hassanaly2023pinn2}}.

\subsection{Battery modeling-specific implementations}

The governing equations can be implemented as described in Sec.~\ref{sec:phyModels} for the physics-loss of the PINN since they involve differentiable functions. The open-circuit potentials $U_{{\rm OCP},j}$ are obtained from experimental observations~\cite{KYCTD22,WKCYGCSGTD23} and need to be converted to differentiable functions. The tool used to convert experimental C/20 $U_{{\rm OCP},j}$ data into differentiable functions is provided in the companion repository. 

\begin{figure}[t!]
	\centering
	\includegraphics[width=65mm]{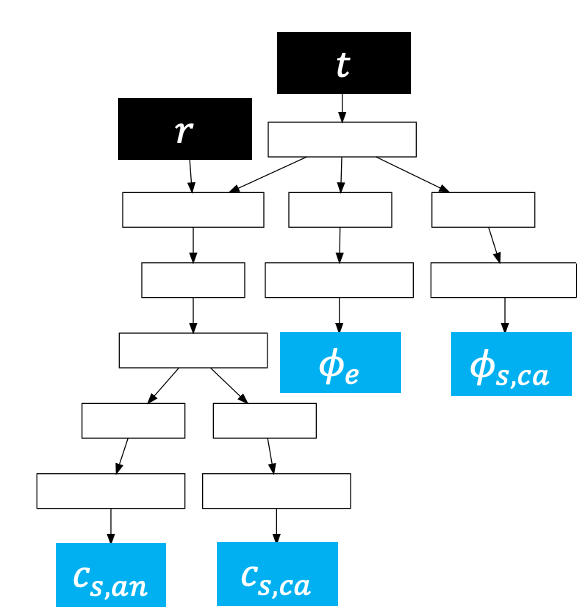}
	\caption{Illustrative PINN architecture used to enforce spatiotemporal dependencies of the state variables for the single-particle model. PINN inputs are in black rectangles, while outputs are in blue rectangles. \revone{White rectangles denote blocks of hidden layers that could be of any type (see Sec.~\ref{sec:arch}).}}
	\label{fig:archspm}
\end{figure}

The neural-net architecture is designed to enforce the appropriate dependencies of the state variables with respect to the spatiotemporal variables as shown in Figure~\ref{fig:archspm}. By construction, while $c_{{\rm s},an}$ and $c_{{\rm s},ca}$ depend on time $t$ and the radial coordinate $r$, the \revtwo{potentials $\phi_{\rm e}$ and $\phi_{\rm s,ca}$} do not depend on the spatial variable $r$. The blank blocks represent the hidden layers and are left blank because they could be of any type without affecting the spatiotemporal dependencies of the state variables. The choice of the hidden layers is further discussed in Sec.~\ref{sec:arch}. When spatial continuity is not needed, the subdomains of the battery are predicted as separate branches to prevent the need to capture unnecessarily large spatial gradients. \revtwo{For example, $\phi_{\rm e}$ and $\phi_{\rm s,ca}$} are predicted by separate branches of the neural net.

Additional constraints are applied via the activation functions used at the output layer. In the case of discharge, the \ce{Li} concentration in the anode must decrease while the \ce{Li} concentration in the cathode must increase over time. In this case, the neural net prediction can be tailored to enforce these monotonic effects. Similar to Eq.~\ref{eq:dist}, in lieu of predicting the physical value of the solid lithium concentration $c_{{\rm s},j}$, an intermediate variable $\widetilde{c_{\rm s}}$ bounded between $0$ and $1$ is predicted. The boundedness is enforced via a sigmoid activation, and $c_{{\rm s},j}(t)$ is reconstructed as 
\begin{equation}
\label{eq:rec_cs}
    c_{{\rm s},j}(t) = \alpha_j \widetilde{c_{\rm s}}(t) F(t) + c_{{\rm s},0,j}.
\end{equation}
In the anode, $\alpha_{\rm an} = - c_{\rm s,0,an}$ and in the cathode, $\alpha_{\rm ca} = c_{\rm s,ca,max} - c_{\rm s,0,ca}$. This approach enforces both the boundedness of $c_{{\rm s},j}$ and their monotonicity over time.

\subsection{Training procedure}
\label{sec:trainingProc}

For the surrogate model training, the input dependent variables (the radial spatial variable $r$ and the time variable $t$) span vastly different scales. While time can span multiple hours, the spatial dimensions are typically on the order of $10^{-5}$~m. Before being passed to the PINN, the input variables are rescaled so they each span the interval of $[0, 1]$. Additionally, the SPM (and P2D model) has potentially coupled, governing equations and boundary conditions for each dependent state variable. Depending on the way the residuals are expressed, the residual equations may have widely different magnitudes. To regularize the residual magnitudes, all residual equations are rescaled by an {\it a priori} estimate of the right-hand side magnitude so that the residual equation magnitude is close to a percentage residual error. 

As is common practice for PINNs~\cite{shin2020convergence,markidis2021old,karniadakis2021physics}, the training procedure first uses batched ADAM~SGD~\cite{kingma2014adam} followed by an L-BFGS full-batch training \cite{fletcher2000practical}. The ADAM part of the training uses a scheduler that decreases the learning rate by an order of magnitude over the last half of the training steps. The L-BFGS uses 50 initial steps to warm-start the approximation of the Hessian. In the cases shown in Sec.~\ref{sec:spmSurr}, the transition to L-BFGS can be subject to instability, which was addressed by using an adaptive learning rate that checkpoints the last gradient descent step and decreases the learning rate if the new loss increased compared to the checkpointed state. As the loss decreases, the learning rate ramps back up to its nominal value. The machine-learning models are implemented using the \verb|tensorflow| library~\cite{tensorflow2015-whitepaper} and the implementation of the PINN is available in the companion repository (\hyperlink{https://github.com/NREL/PINNSTRIPES}{https://github.com/NREL/PINNSTRIPES}).

\section{Single-particle model PINN surrogate}
\label{sec:spmSurr}

\revone{The present section describes the PINN surrogate construction and discusses several strategies to improve the surrogate model for a given computational training budget. A particular focus is given to the PINN surrogate accuracy as compared to a finite-difference physics-based model. The computational speedup obtained by replacing physics-based models with a PINN surrogate is discussed in Part~II \revone{\cite{hassanaly2023pinn2}}. Similarly, the PINN training cost is also reported in Part~II \revone{\cite{hassanaly2023pinn2}}.}

\subsection{Variability with respect to neural network weight initialization}
\label{sec:variability}

\begin{figure}[t!]
	\centering
	\includegraphics[width=0.48\textwidth]{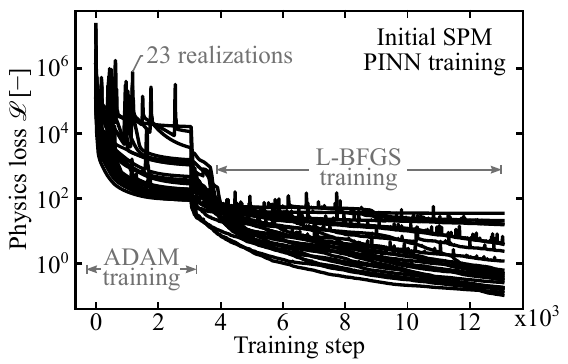}
	\caption{Training loss history for 23 realizations of an SPM PINN surrogate using only the physics loss.}
	\label{fig:var}
\end{figure}

The PINN training process is inherently subject to variability due to the random initialization of the neural network weights and the location of the collocation points. To mitigate this variability, weight initialization is commonly chosen to follow the Glorot normal method (also known as Xavier initialization method)~\cite{glorot2010understanding, markidis2021old}. However, PINNs may still suffer from significant variability across training runs. Figure~\ref{fig:var} illustrates this variability by showing the global loss (as defined by Eq.~\ref{eq:totalLoss}) history of 23 realizations of an SPM PINN surrogate. In this case, each residual is weighted with a coefficient greater than unity (see Sec.~\ref{sec:BalancingPhysicsTermsWithHyperparameters}), which explains the large global loss $\mathcal{L}$ values displayed. The training is repeated 23 times, with each realization using different collocation points $\mathcal{C}$ and weights $\boldsymbol{\theta}$ initialization. Unless otherwise specified, only residual losses $\mathcal{L}_{\rm int} + \mathcal{L}_{\rm bound}$ are minimized, which means that the model is trained without any data, which is similar to the work of Sun et  al.~\cite{sun2020surrogate}. Each case uses a uniform spatiotemporal distribution of 1280 collocation points in the interior of the domain and 640 collocation points at the boundaries. The collocation points are fixed throughout training. During the ADAM training, 10 batches of collocation points are used per epoch. The neural net uses a split architecture (described in Sec.~\ref{sec:arch}) with 1 layer of 20 neurons before the branches and 3 layers constructed as in Wang et al.~\cite{wang2021understanding} with 20 neurons per layer and hyperbolic tangent activation. The learning rate decreases from $10^{-3}$ to $10^{-4}$ during the ADAM training for the first 1500 steps and is held constant for the next 1500 steps. The L-BFGS training is done for 10000 steps. Aside from the architecture, the training details are held fixed in the rest of the manuscript. 

Figure~\ref{fig:var} shows the global loss history with respect to the training steps for 23 realizations of the PINN SPM surrogate.  As illustrated, two prominent features are observed.  First, the two distinct training stages are easily identified with a loss that quickly plateaus during the ADAM training, before again decreasing quickly once the L-BFGS training starts (see plateaus before and after $\approx$3000 steps). Second, there is a large variability between realizations, and this variability persists to the end of training. For each realization, the accuracy of the surrogate PINN is determined by comparing it to a finite-difference PDE solution. The finite difference solution is obtained via implicit Euler integration with a timestep of $0.1$~s and a uniform radial discretization with 64 points for the anode particle (particle radius of $4\mu\rm{m}$) and for the cathode particle (particle radius of $1.8\mu\rm{m}$). The finite-difference solver implementation is available in the companion repository. The scaled mean absolute error $\varepsilon$ for each realization can be expressed as
\begin{equation}
    \label{eq:errSPM}
    \varepsilon = \sum_{\xi \in \{c_{\rm s,an}, c_{\rm s,ca}, \phi_{\rm e}, \phi_{\rm s,ca}\}}  \frac{1}{N_{\xi}} \sum_{i \in [1,N_{\xi}]} \left| \frac{\xi_{\rm PINN, i} - \xi_{\rm PDE, i}}{\xi_{\rm PDE, i} } \right|,  
\end{equation}
where $\xi_{\rm PDE, i}$ is the solution obtained from finite difference at the point $i$, $\xi_{\rm PINN, i}$ is the predicted solution by the PINN surrogate model at the point $i$, and $N_{\xi}$ is the number of points over which the error is computed for each state variable $\xi$. For the training shown in Fig.~\ref{fig:var}, the error varies between $0.02$ at best and $0.39$ at worst. In the following analysis, all surrogate PINN models are trained between 5 and 40 times to account for this initialization variability, and the statistics of the performance are shown instead of only being reported for the best-performing model. Note that if the focus was not on evaluating different training strategies, the variability with respect to the weight initialization could be simply mitigated by training multiple neural nets starting from different initial weights, and choosing the best\revtwothree{-}performing one.



\subsection{Balancing physics terms with penalty parameters}
\label{sec:BalancingPhysicsTermsWithHyperparameters}

Rescaling the governing equation residuals is common in PDE solvers and is often associated with preconditioning \cite{brown1989reduced}. The same problem exists in PINN training, where rescaling residuals solely based on the expected magnitude of the right-hand side might not be sufficient to achieve high accuracy. Here, the residuals are rescaled by coefficients that are optimized via hyperparameter tuning. Note that given the variability observed in Section~\ref{sec:variability}, hyperparameter tuning must reduce the average error given by Eq.~\ref{eq:errSPM} rather than individual realizations of the error. To reduce the number of tunable hyperparameters, the interior residuals of $c_{{\rm s},j}$ are rescaled by the same coefficient $w_{c_{\rm s,int}}$, where the `int' subscript indicates that this acts on dependent variables interior to the domain. The residuals of the boundary conditions of $c_{{\rm s},j}$ at $r=0$ are both rescaled by  $w_{c_{\rm s,rmin}}$ and the residuals of the boundary conditions of $c_{\rm s}$ at $r=R_{{\rm max},j}$ are rescaled by $w_{c_{\rm s,rmax}}$, where `min' and `max' subscripts indicate that these terms act on the respective minimum and maximums of the radial domain, respectively. Therefore, the physics loss at interior collocation points becomes
\begin{equation}
\begin{split}
    \mathcal{L}_{\rm int} &=  w_{c_{\rm s,int}}^2 \bigg( ||{\rm Res}_{c_{\rm s, an, int}} ||_2^2 + || {\rm Res}_{c_{\rm s, ca, int}} ||_2^2\bigg) \\&~~+ || {\rm Res}_{\phi_{\rm e, int}} ||_2^2  + || {\rm Res}_{\phi_{\rm s,ca, int}} ||_2^2,
\end{split}
\end{equation}
where $|| x ||_2^2 = \frac{1}{N_{\mathcal{C}}}\sum_{i \in \mathcal{C}} x_i^2$, ${\rm Res}_{c_{\rm s, an, int}}$ (resp. ${\rm Res}_{c_{\rm s, ca, int}}$) is the residual of the \ce{Li} solid concentration in the anode (resp. cathode) rescaled by typical value, ${\rm Res}_{\phi_{\rm e, int}}$ is the residual of the potential in the electrolyte rescaled by its typical value, and ${\rm Res}_{\phi_{\rm s, ca, int}}$ is the residual of the potential in the cathode rescaled by its typical value.

Likewise, the boundary loss becomes
\begin{equation}
\begin{split}
    \mathcal{L}_{\rm bound} &=   w_{c_{s,rmin}}^2 \bigg(|| {\rm Res}_{c_{\rm s, an, rmin}} ||_2^2 + || {\rm Res}_{c_{\rm s, ca, rmin}} ||_2^2\bigg)  \\&~~+ w_{c_{\rm s,rmax}} ^2 \bigg(|| {\rm Res}_{c_{\rm s, an, rmax}} ||_2^2 + || {\rm Res}_{c_{\rm s, ca, rmax}} ||_2^2\bigg),
\end{split}
\end{equation}
where ${\rm Res}_{c_{\rm s, an, rmin}}$ (resp. ${\rm Res}_{c_{\rm s, ca, rmin}}$) is the residual of the \ce{Li} solid concentration boundary condition at $r=0$ in the anode (resp. cathode), and ${\rm Res}_{c_{\rm s, an, rmax}}$ (resp. ${\rm Res}_{c_{\rm s, ca, rmax}}$) is the residual of the \ce{Li} solid concentration boundary condition at $r=R_{{\rm max,an}}$ in the anode (resp. cathode).

\begin{figure*}
    \centering
    \includegraphics[width=100mm]{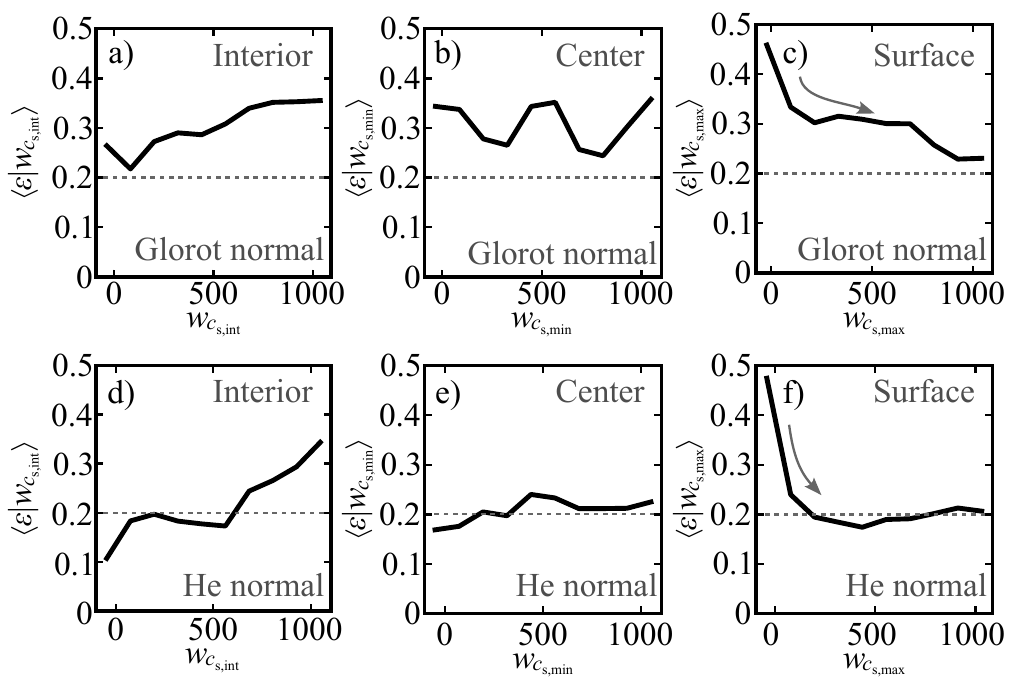}
    \caption{Conditional average of the PINN error conditioned on the weight $w_{c_{\rm s,int}}$ for the collocation points located in the interior of the SPM spatial domain (a, d), the collocation points located at the $r=0$ boundary (b, e) and the collocation points located at the $r=R_{{\rm max},j}$ boundary (c, f). Weight initialization is Glorot normal (a, b, c) and He normal (d, e, f).}
	\label{fig:gpopt}
\end{figure*}

The global loss, in the absence of data, only includes the interior and boundary loss, i.e., Eq.~\ref{eq:totalLoss} becomes
\begin{equation}
    \mathcal{L} =  \mathcal{L}_{\rm int}  +  \mathcal{L}_{\rm bound}.
\end{equation}
In total, 150 runs were simulated with random sampling of the physics loss weights spanning the range of $[0.1, 1000]$. The parameter sweep is simulated with two different weight initialization techniques: the Glorot initialization \cite{glorot2010understanding} and the He initialization \cite{he2015delving}. The accuracy of each neural net is then evaluated based on the error defined in Eq.~\ref{eq:errSPM}. 

Figure~\ref{fig:gpopt} shows the average loss over all the training conditioned on each one of the weight values. On average the He normalization provided higher accuracy than the Glorot normalization (see dashed lines in Fig.~\ref{fig:gpopt}). The sensitivity of the error with respect to the loss weights follows similar trends for both initializations. Low values of $w_{c_{\rm s,int}}$ and high values of $w_{c_{\rm s,rmax}}$ lead to higher accuracy. This tendency can be physically interpreted as the diffusion in the spherical particles is driven by the concentration gradient at the particle surface. In practice, a common failure mode of the PINN is to predict the trivial solution (i.e., no change in time) for the particle diffusion process, which achieves low interior residuals but high boundary residuals. This effect is even more prominent for the P2D model discussed in Part~II \revone{\cite{hassanaly2023pinn2}}. An increased emphasis on the particle boundary conditions appears to successfully avoid this failure mode. The PINN accuracy appears to be almost independent of the boundary condition enforcement at $r=0$ (see center plots in Fig.~\ref{fig:gpopt}). In practice, this boundary condition is easily satisfied even with non-fully trained PINNs. In the rest of the manuscript, the results of this analysis are used to set the weights of the residuals. The variability in the accuracy was also evaluated throughout the hyperparameter space (not shown here) and was not as heavily impacted by the choice of the loss weights as the mean accuracy.

\subsection{Architecture model effect}
\label{sec:arch}

\begin{figure}[th!]
     \centering
     \begin{subfigure}[b]{0.49\textwidth}
         \centering
         \includegraphics[width=\textwidth]{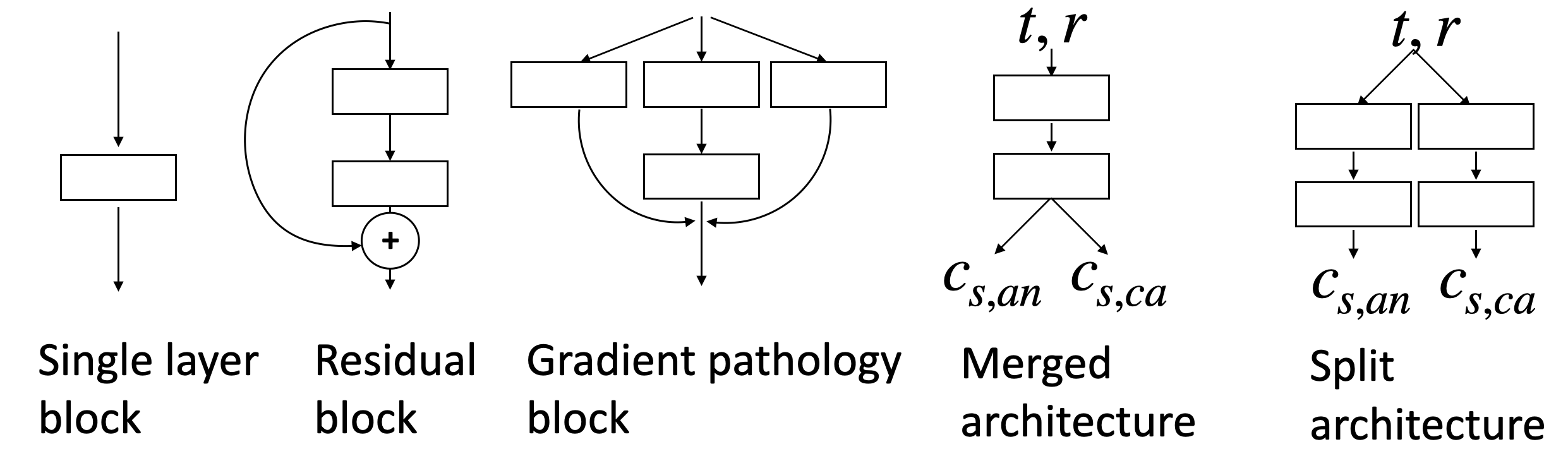}
         \caption{ }
         \label{fig:arch_blocks}
     \end{subfigure}
     \begin{subfigure}[b]{0.49\textwidth}
         \centering
         \includegraphics[width=3.191in]{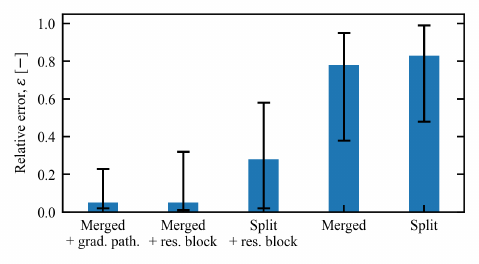}  
         \caption{ }
         \label{fig:arch_spm_comp}
     \end{subfigure}
        \caption{Comparison of neural net architectures for the PINN surrogate. (a) Schematic representation of the neural net architectures implemented. (b) Average PINN error over all the training realizations (bar height) for different neural network architectures. The error bar denotes the 95\% percentile variability observed for all the realizations.}
        \label{fig:sec_cons}
\end{figure}

The choice of the neural net architecture has been demonstrated to be important to alleviate typical pathologies in PINNs \cite{wang2021understanding}. In the present section, different architectures are compared to identify which architecture leads to the best accuracy, and are shown in Fig.~\ref{fig:arch_blocks}. The first architecture proposed is the \textit{split} architecture where the four state variables $\{ c_{\rm s,an}, c_{\rm s,ca}, \phi_{\rm e}, \phi_{\rm s,ca}\}$ are predicted independently from one another by each branch of the network. These branches are then coupled together via the loss function (Eq.~\ref{eq:totalLoss}). The advantage of this architecture is that the weights are not shared across variables which allows the branches to be specialized in predicting specific state variables. The second architecture chosen is the \textit{merged} architecture where the spatio-temporal variables are first transformed by several layers before branching out into the four state variables. The advantage of this configuration is that since the change of one variable is expected to be coupled to the other ones, one would avoid encoding the same information multiple times. For both the \textit{merged} and the \textit{split} approach, one can also replace the standard layers with residual blocks which have been successful in a variety of applications~\cite{he2016deep}. Finally, a specific architecture of blocks is described in Ref.~\cite{wang2021understanding}, referred to as \textit{gradient pathology}, which combines residual blocks and multiplicative coupling is tested. For all the architectures, the number of layers is adjusted to ensure that the networks all have $\approx$9000 trainable parameters. 

Assuming $\alpha_{\rm a} = 0.5$, the Butler--Volmer reaction (Eq.~\ref{eq:BV}) can be linearized as  
\begin{equation}
\label{eq:bvLin}
\begin{split}
    J_j = i_{0,j}\frac{\phi_{{\rm s},j}-\phi_{\rm e}-U_{{\rm OCP},j}}{RT}.
\end{split}
\end{equation}
Unlike the fully non-linear Butler--Volmer formulation, the linearized Butler--Volmer formulation typically prevents very small or very large gradients in the presence of inaccuracies in the potentials. This simplification is used in the present architecture comparison study.  The original non-linear Butler--Volmer reaction is reintroduced later in Sec.~\ref{sec:NonlinBV}.  The models are evaluated against the solution of the PDEs described in Sec.~\ref{sec:phyModels}. Figure~\ref{fig:arch_spm_comp} shows the PINN error $\varepsilon$ (Eq.~\ref{eq:errSPM}) using different combinations of PINN architectures. A primary observation is that the residual blocks or the gradient pathology blocks led to PINNs with lower errors as compared to PINNs developed with different architectures. The gradient pathology blocks appear to be slightly superior as they provide less variability in the PINN performances. Next, the split architecture is consistently outperformed by the merged architecture which suggests that enforcing a tight coupling between the variables is necessary for the SPM. \revone{Additional comments on the architecture effects are provided in Sec.~\ref{sect:Discussion}.}

%

\subsection{PINN training regularization}
\label{sec:trainreg}

Several training regularization strategies have been proposed to encourage the residual minimization step to consistently lead to improved accuracy. In the present section, multiple training regularizations are adopted to attempt to further improve the PINN accuracy. All the regularization methods use the merged architecture with gradient pathology blocks (the best-performing architecture in Sec.~\ref{sec:arch}). 

Recently, a sequence-to-sequence learning regularization method (based on slowly increasing the time interval covered by the collocation points) has been found o successfully address training instabilities, specifically for 1D reaction-diffusion PDEs \cite{krishnapriyan2021characterizing}. Here, this approach is implemented in two ways: 1) during the ADAM training, the temporal domain over which the collocation points are sampled is stretched at every epoch during the first half of the epochs (number of times the entire training set is shown to the network). During the later half of the epochs and the L-BFGS training, the temporal domain is held fixed at its maximal extent. This approach is referred to as \textit{Gradual SGD}; 2) During the ADAM training, the full extent of the temporal domain is shown to the network. During the L-BFGS training, the temporal domain is slowly increased. This approach is referred to as \textit{Gradual L-BFGS}. 

The placement of the collocation points can also be altered to improve generalizability. Instead of choosing a set of collocation points for the entire training procedure, the collocation points are randomly resampled at every epoch during the ADAM training. During the L-BFGS training, the collocation points are then held fixed. This approach is referred to as \textit{Random collocation}. 

The PINN training can be viewed through the same lens as multi-task learning in computer vision where loss weighting is paramount \cite{kendall2018multi}. Several training regularization methods have been introduced to strategically weights the residuals in the loss function. A recent regularization technique based on attention mechanisms has been proposed to emphasize parts of the spatio-temporal domain where residuals are especially difficult to capture~\cite{mcclenny2020self}. In this technique, every collocation point is assigned a weight that is trained during the ADAM training procedure. To avoid biasing the attention based on the error made in the initial training stages, the weights are trained only after the first half of the epochs are finished during the ADAM training stage. During the L-BFGS training, the weights are held fixed. This approach is referred to as \textit{self-attention}.

Finally, a gradient annealing procedure proposed by Wang et al.\cite{wang2021understanding} is used to balance the physics losses. The method adjusts the weight of each physics loss term to balance the magnitude of the gradient induced by each term. The gradient annealing method uses a moving average factor of $0.9$, consistently with Wang et al.~\cite{wang2021understanding}.

Similar to Sec.~\ref{sec:arch}, the regularization methods are evaluated based on the value of $\varepsilon$ (Eq.~\ref{eq:errSPM}).  Perhaps disappointingly, none of the regularization techniques described above exceeded the performance of the base method. Additionally, the variability in the training results was found to be lowest without any PINN-specific regularization. While these results are statistically significant, they could be explained by the specific equations adopted or by the fact that the regularization was implicitly already implemented by weighting the residuals (in Sec.~\ref{sec:BalancingPhysicsTermsWithHyperparameters}).   

\begin{figure}[th!]
	\centering
        \includegraphics[width=3.176in]{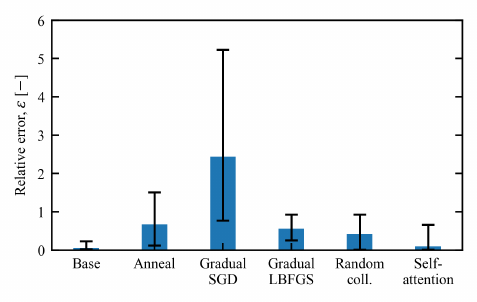}

	\caption{Average PINN error (bar height) for different training regularization strategies. The error bar denotes the 95\% percentile variability observed for all the realizations.}
	\label{fig:reg_spm_comp}
\end{figure} 

\begin{figure}[th!]
     \centering
     \begin{subfigure}[b]{0.4\textwidth}
         \centering
         \includegraphics[width=2.202in]{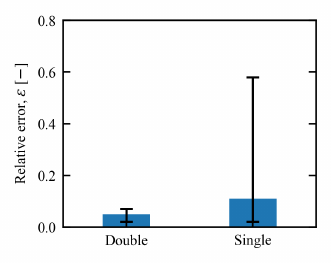}
         \caption{}
         \label{fig:prec_spm}
     \end{subfigure}
     \hfill
     \begin{subfigure}[b]{0.4\textwidth}
         \centering
         \includegraphics[width=2.202in]{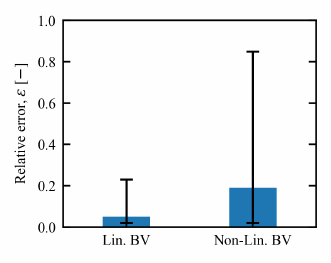}
         \caption{}
         \label{fig:comp_NLBV}
     \end{subfigure}
        \caption{Average PINN error (bar height). The error bar denotes the 95\% percentile variability observed for all the realizations effect of float precision. (a) Effect of float precision on the PINN error. (b) Effect of the linearization of the Butler--Volmer formulation on the PINN error.}
        \label{fig:prec_spm_comp_NLBV}
\end{figure}

As a complementary computational experiment, the effect of precision was also evaluated by either using double precision or single precision. Using lower precision is typically advantageous to reduce the memory pressure on the devices used to train the model and reduce the computational cost of training. Figure~\ref{fig:prec_spm_comp_NLBV}a compares the accuracy obtained when using double-precision and single-precision floats.  Consistent with prior observations, the use of double precision improves the PINN accuracy \cite{mao2020physics} (here by a factor 2). It is also found that it reduces the training variability. \revtwotwo{Although the effect of training regularization methods and hyperparameter choices was demonstrated on a 2C discharge case only, it is expected that the same training procedure would perform similarly with other current conditions. Appendix~\ref{sect:SPM_otherCC_Appendix} shows that the PINN accuracy at other constant current conditions (constant or time-varying) is on par with the 2C rate used throughout Sec.~\ref{sec:spmSurr}.}

\begin{figure}[th!]
	\centering
    \includegraphics[width=2.702in]{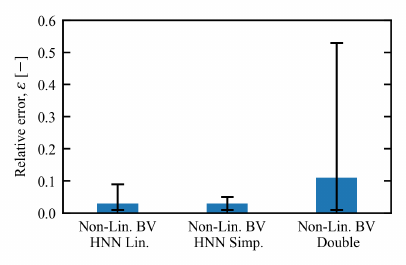}
	\caption{Average PINN error (bar height) for hierarchical training and non-hierarchical training with the same parametric expressivity. The error bar denotes the 95\% percentile variability observed for all the realizations.}
	\label{fig:NLBVHNN}
\end{figure} 

\begin{figure}
    \centering
    \includegraphics[width=78mm]{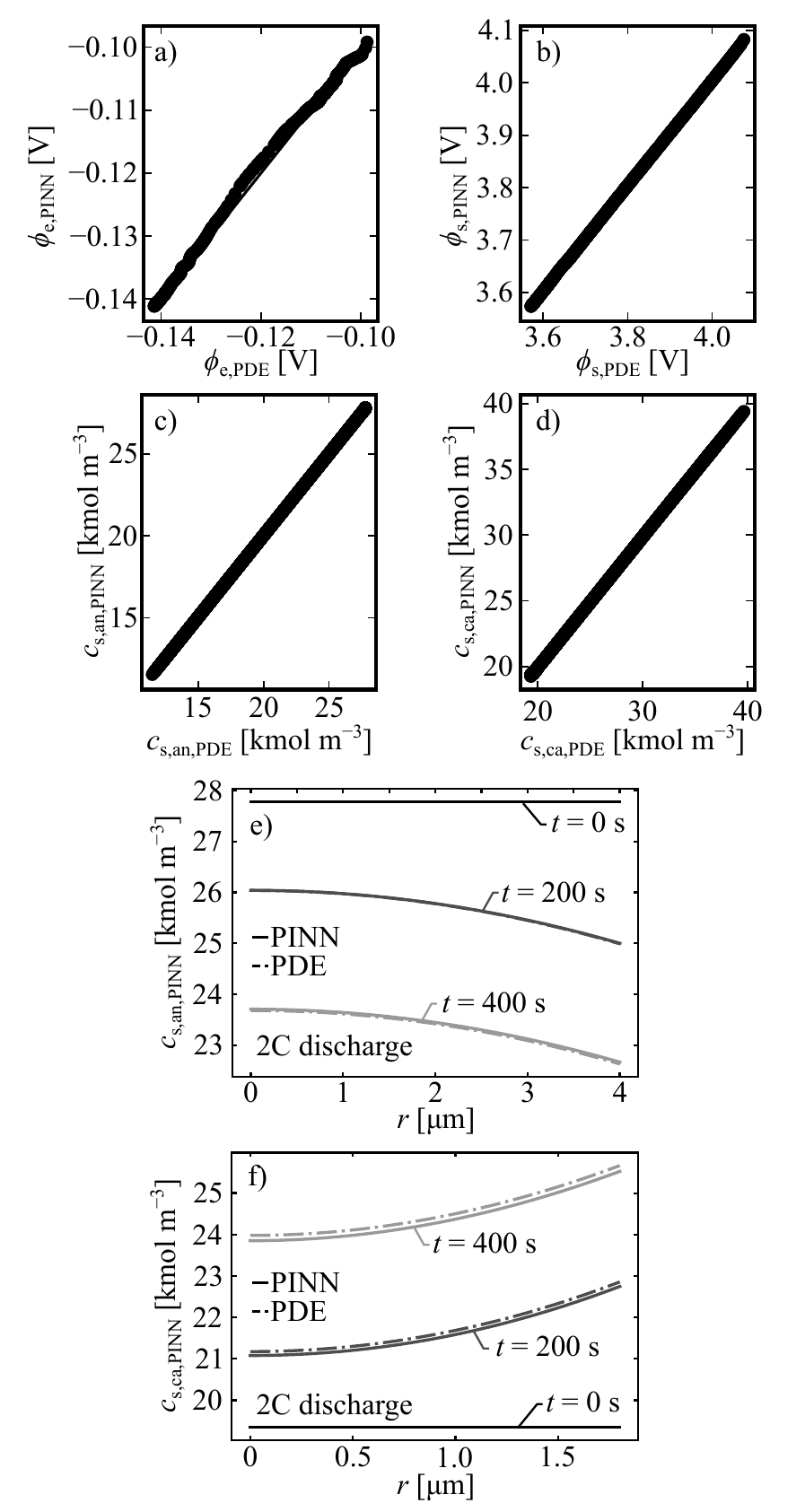}
    \caption{\revtwo{(a-d) 45$^\circ$ correlation plot between state variables predicted via PDE integration ($x$-axis) and predicted with the PINN ($y$-axis). (e-f) Solid-phase \ce{Li} concentration with respect to radius at 0~s (\mythickline{black}), 200~s (\mythickline{darkgray}) and 400~s (\mythickline{gray}), into a 2~C discharge for the anode and cathode, for the PINN prediction (\mythickline{black}) and the PDE solution (\mythickdasheddottedline{black}).}}
    \label{fig:spm_res}
\end{figure}

\subsection{Fully non-linear model and training hierarchy}
\label{sec:NonlinBV}

As mentioned in Sec.~\ref{sec:arch}, the SPM equations implemented in the PINN so far were simplified by linearizing the Butler--Volmer formulation.  To use the original, non-linear Butler--Volmer expression, the PINN can be trained to handle reaction nonlinearities. In practice, the non-linearity can lead to training instabilities due to the exponential increase of the loss function with respect to errors in the predicted potentials. This issue can be addressed by gradient clipping or by clipping the Butler--Volmer reaction term, which can lead to the opposite problem where the gradients passed to the PINN are too small. Figure~\ref{fig:prec_spm_comp_NLBV}b shows the error $\varepsilon$ when training the PINN with a linear Butler--Volmer reaction as compared to training the PINN with a non-linear Butler--Volmer reaction. It is clear that while the non-linear version can lead to reasonable accuracy, it can also catastrophically fail and lead to average errors that are larger than the errors induced when training with a linear Butler--Volmer reaction by a factor 5.   

To handle the inherent instability of the non-linear Butler--Volmer during training, we propose a remedy based on a hierarchy of PINNs inspired by curriculum learning regularization~\cite{krishnapriyan2021characterizing}. In curriculum learning, instead of solving the full non-linear problem all at once, the solution may be approximated by a sequence of neural nets trained with increasingly higher fidelity. In the present context, fidelity refers to the fidelity of the physics-loss minimized. The first fidelity level may be the SPM governing equations that use a linear Butler--Volmer reaction. The second fidelity level may then be the SPM solution that uses a non-linear Butler--Volmer reaction. The overall objective is to approximate the solution of the physics loss at the second fidelity level. In the multi-fidelity approach, one first approximates the solution of the lower-fidelity governing equations and then learns to correct the initial approximation to obtain the solution of the higher-fidelity governing equations. This approach is typical of bi-fidelity modeling~\cite{DBRSJA20,DD22,de2023bi} and multi-stage training~\cite{hassanaly2023uniform}, where learning to correct a lower fidelity model can be easier than learning the high-fidelity model directly. \revtwothree{Within the battery modeling context, the proposed hierarchical training echoes the hybrid physics-based and data-based modeling framework, where a data-based model is trained to correct predictions of an erroneous physics-based model \cite{tu2023integrating}.} In the present case, one neural network predicts $\widetilde{\xi}_{1}(t)$ constrained by residuals defined with a linear Butler--Volmer relation, which is used to reconstruct $\xi_{1}(t)$ as
\begin{equation}
    \xi_{1}(t) = \widetilde{\xi}_{1}(t) F(t) + \xi_{0}.
\end{equation}
The solution predicted by the first neural net is frozen and a second neural network is trained to predict $\widetilde{\xi}_{2}(t)$ which is constrained by the residuals defined with a non-linear Butler--Volmer relation and is used to reconstruct $\xi_{2}(t)$ as
\begin{equation}
    \xi_{2}(t) = \bigg(\alpha_2 \widetilde{\xi}_{2}(t) + \xi_{1}(t) - \xi_{0}\bigg) F(t) + \xi_{0}.
\end{equation}
$\alpha_2$ is a coefficient typically lower than unity that incentivizes the second-level to correct the first level with a low-amplitude signal. Here, $\xi_{2}(t)$ is reconstructed differently than $\xi_{1}(t)$ to 1) force the second level to start from the solution $\xi_{1}(t)$ thereby preventing landing other local minima of the loss function;  2) to prevent $\widetilde{\xi}_{2}(t)$ from capturing the spatiotemporal features already captured by $\xi_{1}(t)$, thereby focusing the expressiveness of the PINN at the second level on correcting the lower-fidelity solution. Therefore, the procedure implicitly assumes that the first fidelity level is already a good approximation of the next fidelity level. If this is not the case, then capturing the correction across fidelities would not be easier and could be even be more complex than capturing the higher fidelity directly. In our procedure, each member of the multi-fidelity hierarchy is trained with a different set of collocation points in order to encourage the final model to minimize residuals over the entire spatiotemporal domain. The effect of this choice is evaluated hereafter.

The multi-fidelity training approach can be deployed with any number of levels. Note, however, that for every level added, the loss function calculation requires the evaluation of all the neural networks in the training hierarchy, which can become costly. In Part II \revone{\cite{hassanaly2023pinn2}}, a hierarchy made of 4 levels is showcased. This approach can also be used with any type of multi-fidelity hierarchy that is sufficiently correlated with the higher fidelity target. Here, the hierarchical training uses a first level obtained from training the PINN with a linear Butler--Volmer relation (referred to as \textit{HNN Lin.}) and an even simpler PINN that uses a linear Butler--Volmer relations, constant diffusivities and linear $U_{{\rm ocp},j}$ (referred to as \textit{HNN Simp.}). To illustrate the benefit of the hierarchical approach, the results are benchmarked against the same PINN as shown in Fig.~\ref{fig:NLBVHNN} except that the number of layers and the number of epochs are doubled (referred to as \textit{Double}) to be appropriately compared to the sequence of networks involved in the hierarchies. Note that the computational cost incurred by the latter strategy is in practice about twice as large as for the hierarchical PINNs since twice as many weights are trained at the same time.

Figure~\ref{fig:NLBVHNN} shows the accuracy of the models trained with the proposed hierarchical approach. As illustrated, using a hierarchical architecture significantly improves the predictions and dramatically reduces the variability in the predictions. On average, the \textit{HNN Simp.} and the \textit{HNN Lin.} approaches led to similar results, which suggests that there are several possible hierarchy choices that result in reduced errors.

Figure~\ref{fig:spm_res} illustrates the PINN SPM surrogate predictions using the \emph{Non-Lin.~BV HNN Lin.}~hierarchy architecture during a 2~C discharge. Figure~\ref{fig:spm_res}a-d illustrate the dependent variables \revtwo{$\{ c_{\rm s,ca},\phi_{\rm s,ca},~c_{\rm s,an}~\phi_{\rm e}\}$} predicted by the PINN surrogate model ($y$-axis) as compared to the dependent variables predicted using the finite-difference PDE solver. In these plots, a 45$^\circ$ line indicates that the two models produce the same dependent variables across their respective spatiotemporal domains.  Figure~\ref{fig:spm_res}e-f illustrates these same dependent variables predicted by the PINN SPM surrogate in a method more consistent with battery literature~\cite{DWZCBJK18,GSW11,SGRW06,MAKMK17}.  
\revtwo{Figure~\ref{fig:spm_res}e and Fig.~\ref{fig:spm_res}f illustrate the anode and cathode solid phase concentration with respect to the radial coordinate, respectively.  Without using a data loss $\mathcal{L}_{\rm data}$, the predicted values of the state variables agree well with the ones obtained with the PDE integration (also shown in Fig.~\ref{fig:spm_res}e-f)}, which indicates that the PINN SPM surrogate can be trained effectively with just using physics loss (i.e., the residuals of the governing equations).

\revall{
The neural net architecture of the best\revtwothree{-}performing case ( \emph{Non-Lin.~BV HNN Lin.}) is summarized in Tab.~\ref{tab:architecture}. A merged architecture is using 1 fully connected (denoted as FC) layer with 20 neurons in the merged part of the network that connects to the time variable $t$, and in the merged part of the network that connect\revtwothree{s} to the radial spatial variable $r$. The branch part of the network (that connects the merged part to the predicted dependent variables, uses 3 gradient pathology (denoted as GP) blocks with 20 neurons per layer as well. The activation of the final layer is linear for the potentials $\phi_{\rm s,ca}$ and $\phi_{\rm e}$ and is sigmoidal for $c_{\rm s, an}$ and $c_{\rm s, ca}$ (consistently with Eq.~\ref{eq:rec_cs}). The first and the second level\revtwothree{s} of the hierarchy use the same architecture.} 

\begin{table}[!hbt]
    \centering
    \revall{\begin{tabular}{c|l|l}
          & First level & Second Level \\
         \hline
         Neuron per layer & 20 & 20  \\
         Activation& $\operatorname{tanh}$ & $\operatorname{tanh}$ \\
         Merged part & 1 (FC) & 1(FC)  \\
         Branch part & 3 (GP) & 3 (GP) \\      
    \end{tabular}}
    \caption{\revall{\emph{Non-Lin.~BV HNN Lin.} model architecture.}}
    \label{tab:architecture}
\end{table}


\section{Discussion}
\label{sect:Discussion}

Designing a surrogate model for parametric inference poses a question of data efficiency. On the one hand, the model must be able to ingest large amounts of data that describe the battery state dynamics for cases where the PDE solution is available. On the other hand, the data-driven surrogate model must cope with low data availability because the PDE solution data becomes intractable for the high-dimensional parameter space. This unique data-availability regime can be addressed via PINNs, which use data where available and the governing equations of the underlying physical system elsewhere. However, successfully training PINNs can be challenging and while various regularizations have been developed they are not necessarily beneficial for battery models. 

This work demonstrates how to best handle the zero-data availability limit without significantly increasing the computational cost by deriving guiding principles for the design of the PINN SPM surrogate using architecture and regularization. For example, spatiotemporal dependence and independence can be strictly enforced via neural network architecture choices and explicit separation of the variables over the domain of definition, thereby allowing for sharp discontinuities (Sec.~\ref{sec:models}). The success of the merged neural network architecture (Sec.~\ref{sec:arch}) \revone{suggests that it is preferable for the PINN to encode the global battery dynamics, rather than specializing in predicting each state variable separately. This finding echo\revtwothree{e}s one observed on visualization tasks where multi-task prediction (say, image depth and segmentation) is more accurate as compared to single-task prediction~\cite{kendall2018multi}. In visualization tasks, the improved accuracy was attributed to the fact that the visualization tasks require learning similar features. In the present case, each ``task'' is predicting each one of the battery state variables. The exact reason behind the improved accuracy of the merged architecture is unclear. Either, it could be that describing the variation of state variable with a common latent space is parameter-efficient, or it could be that the construction of a common latent space improves the PINN loss landscape and prevents landing on a local minimum. As future work, it would also be interesting to explore which variables are more appropriate to couple with one another (i.e., $c_{\rm s,an}$ and $c_{\rm s,ca}$ likely follow more similar dynamics as compared to $c_{\rm s,an}$ and $\phi_{\rm e}$).} 

Accurate surrogate model predictions can also be obtained by appropriately weighting the loss residuals, which can be done {\it a priori}. In training, it was found that the solid-surface flux due to electrochemical reactions must be appropriately weighted as compared to the internal conservation of species equation residuals to avoid the trivial solution (Sec.~\ref{sec:BalancingPhysicsTermsWithHyperparameters}). Despite the relatively simple description of Li-ion battery physics in the SPM, several training difficulties were identified. First, the relative weights of the physics residuals need to be appropriately chosen to emphasize that the dynamics are driven by the particle surface boundary conditions. Second, significant training variability could be observed, and high accuracy could be challenging to achieve, especially because of the non-linearity in the Butler--Volmer formulation (Sec.~\ref{sec:NonlinBV}). These difficulties are found to be effectively addressed by using a curriculum learning approach where the complexity of the governing equation is gradually increased. Here a multi-fidelity training hierarchy is proposed and shown to be successful. Other approaches such as transfer learning could also be envisioned but are left for future work. The PINN design guiding principles described in this work will be applied in Part~II \revone{\cite{hassanaly2023pinn2}} for the P2D model and allow for efficient parameter inference. 

\section{Conclusion}
\label{sect:Conclusion}
The present manuscript describes a method for implementing a physics-informed neural network as a surrogate of a single particle model for \ce{Li}-ion batteries.  On initial implementation, the PINN surrogate suffered from significant variability and was, in some instances, unable to produce reasonable accuracy.  To improve the PINN accuracy, several approaches such as hyperparameter balancing, architecture effects, regularization techniques, and hierarchical training were considered.  In this study, it was found that using a neural network architecture that leverages the coupling between state variables, and that using residual blocks can significantly improve the performance of a PINN when no data (i.e., solutions to the governing equation set) is available.  It was also shown that linearizing the Butler--Volmer reaction can improve the PINN surrogate performance. In case one would prefer to not to linearize the reaction kinetics, a hierarchical training approach was shown to lead to reliable and accurate results, even in the presence of non-linear reaction kinetics. Given the similarity of the SPM equation with the higher fidelity pseudo-two-dimensional (P2D) equations, some of the approaches taken in the present work to develop a PINN SPM surrogate are applied to developing PINN P2D surrogate model in Part~II \revone{\cite{hassanaly2023pinn2}}.

\section*{Acknowledgements}
This work was authored by the National Renewable Energy Laboratory (NREL), operated by Alliance for Sustainable Energy, LLC, for the U.S. Department of Energy (DOE) under Contract No. DE-AC36-08GO28308. This work by authored in part by Idaho National Laboratory (INL) operated by Battelle Energy Alliance, LLC under contract No. DE-AC07-05ID14517. This work was supported by funding from DOE's Vehicle Technologies Office (VTO) with Simon Thompson as program manager and DOE's Advanced Scientific Computing Research (ASCR) program with Steven Lee as program manager. A.D.'s work was also partially supported by the AFOSR awards FA9550-20-1-0138. The research was performed using computational resources sponsored by the Department of Energy's Office of Energy Efficiency and Renewable Energy and located at the National Renewable Energy Laboratory. The views expressed in the article do not necessarily represent the views of the DOE or the U.S. Government. The U.S. Government retains and the publisher, by accepting the article for publication, acknowledges that the U.S. Government retains a nonexclusive, paid-up, irrevocable, worldwide license to publish or reproduce the published form of this work, or allow others to do so, for U.S. Government purposes.

\section*{Nomenclature}
\scriptsize
\begin{tabbing}
	Variable \hspace{5mm}\= Description \hspace{38mm} \= SI Units \\
        $c_{\rm e}$             \> Electrolyte Li-ion concentration \> kmol~m$^{-3}$\\
        $c_{{\rm s},j}$         \> Solid-phase Li concentration in phase $j$\>  kmol~m$^{-3}$\\
        $\widetilde{c}_{{\rm s}}$  \> Normalized solid-phase Li concentration \>  $-$\\
        $c_{{\rm s},0,j}$       \> Initial solid-phase Li concentration in phase $j$\> kmol~m$^{-3}$\\
        $c_{{\rm s, max},j}$    \> Max solid-phase Li concentration in phase $j$\> kmol~m$^{-3}$\\
        $\boldsymbol{d}$            \> Experimental observations \>  \\
        $D$                     \> Ramping function \> \\
        $D_{{\rm s},j}$         \> Solid-phase Li diffusion coefficient \> m$^2$~s$^{-1}$\\
        $F$                     \> Faraday's constant \> s~A~kmol$^{-1}$\\
        $i_0$                   \> Exchange current density \> A~m$^{-2}$ \\
        $i^0_0$                 \> Exchange current density prefactor \> A~m$^{-2}$ \\
        $I$                     \> Current demand \> A~m$^{-2}$\\
        $j$                     \> Phase indicator \> $-$ \\
        $J_j$                   \> Li-ion flux due to electrochemical reactions \> kmol~m$^{-2}$~s$^{-1}$\\
        $\mathcal{L}$           \> Global loss \> $-$ \\
        $\mathcal{L}_{\rm int}$ \> Interior collocation point residual avg. error \\
        $\mathcal{L}_{\rm bound}$ \> Spatial boundary collocation point residual avg. error \\
        $\mathcal{L}_{\rm data}$ \> Avg. error against available data \\
        $\boldsymbol{p}$            \> Parameter set \>  \\
        $p_{\rm like}$          \> Likelihood function \> $-$ \\
        $p_{\rm post}$          \> Posterior probability function \> $-$ \\
	$p_{\rm prior}$         \> Prior probability function \> $-$ \\
        $r$                     \> Radial coordinate \> m \\
        $R$                     \> Universal gas constant \> J~kmol$^{-1}$~K$^{-1}$\\
        Res                     \> Residual of governing equation   \> \\
        $t$                     \> Time \> s \\
        $T$                     \> Temperature \> K \\
        $U_{{\rm OCP},j}$       \> Open-circuit potential of active material $j$ \> V\\
        $V_j$                   \> Total volume of composite $j$ \> m$^3$\\
        $\alpha_2$              \> Hierarchical scale factor for second-level PINN \> $-$\\
        $\alpha_{\rm a}$        \> Anodic symmetry factor \> $-$ \\
        $\alpha_{j}$            \> Phase concentration scaling factor \> kmol~m$^{-3}$\\
        $\epsilon_j$            \> Active material volume fraction in phase $j$ \> $-$\\ 
        $\varepsilon$           \> Scaled mean absolute error \> $-$ \\ 
        $\xi$                   \> Predicted state variable \> \\
        $\xi_{\rm m}$           \> Predicted state variable from model $m$\> \\
        $\xi_0$                 \> Initial condition of state variable \> \\
        $\xi_1$                 \> PINN predictions using linear Butler--Volmer \> \\
        $\xi_2$                 \> PINN predictions using Butler--Volmer \> \\
        $\widetilde{\xi}$       \> Raw predicted state variable from neural net\> \\
        $\tau$                  \> Timescale with significant init.~condition effect \> s\\
        $\phi_{\rm e}$          \> Electrolyte potential \> V\\
        $\phi_{{\rm s},j}$      \> Solid-phase potential in composite $j$\> V \\
\end{tabbing}

\bibliographystyle{elsarticle-num}

\section*{Appendix}
\appendix
\renewcommand{\thesection}{\Alph{section}}
\setcounter{section}{0}
\setcounter{figure}{0}
%
\small

\section{\revtwothree{Applicability to other constant-current and smoothly varying current conditions}}
\label{sect:SPM_otherCC_Appendix}

\begin{figure*}[th!]
	\centering
        \includegraphics[width=3.176in]{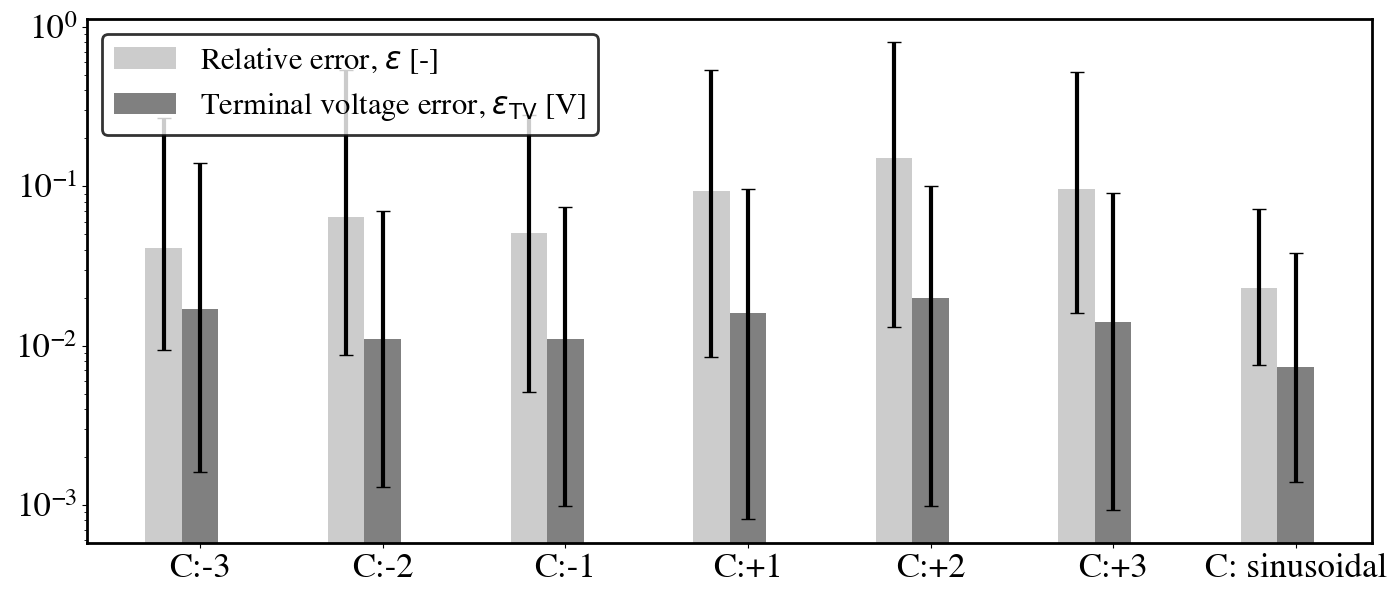}
	\caption{\revtwotwo{Average PINN relative error $\varepsilon$ (darker bar) and terminal voltage error $\varepsilon_{\rm TV}$ computed with Eq.~\ref{eq:errtv} (lighter bar) for different current conditions. The error bar denotes the 95\% percentile variability observed for all the realizations.}}
	\label{fig:spm_cc}
\end{figure*} 

\revtwotwo{The training choices related to balancing physics losses (Sec.~\ref{sec:BalancingPhysicsTermsWithHyperparameters}), network architecture (Sec.~\ref{sec:arch}) and training regularization (Sec.~\ref{sec:trainreg}) were all investigated with a 2C rate discharge. To verify that the results also hold to different current conditions, the same architecture shown for the first level of Tab.~\ref{tab:architecture}, and the same loss weights are used to train a PINN that predicts the solution of the SPM with different current conditions. Six different PINN surrogates are trained with six different C-rates: three discharging rates denoted as $\{ C:-3 , C:-2, C:-1\}$, and three charging cases $\{C:+1, C:+2, C:+3\}$. The integer value denotes the C-rate and the sign denotes the charge/discharge direction. To compare the different cases, the time interval over which the PINN is trained is adapted to the C-rate. For C-rate of $1$ (charge or discharge), the total time interval spanned by the collocation points is $[0, 2700~{\rm s}]$, for a C-rate of $2$, it is $[0, 1350~{\rm s}]$ and for a C-rate of $3$ it is of $[0, 900~{\rm s}]$. The charging cases only differ from the discharge in the way monotonicity is enforced for the solid concentration (see Eq.~\ref{eq:rec_cs}). Compared to the discharge case, Li concentration in the anode is constrained to only increase and Li concentration in the cathode is constrained to only decrease. This is achieved by still rescaling the solid Li concentration using Eq.~\ref{eq:rec_cs}, but by using $\alpha_{\rm an} = c_{\rm s,an,max} - c_{\rm s,0,an}$ and in the cathode, $\alpha_{\rm ca} = - c_{\rm s,0,ca}$. All the other training parameters are held constant and all the cases use a linear Butler--Volmer relation. The models are trained using only a physics loss during 3000 ADAM SGD epochs and 20000 L-BFGS epochs.}

\revtwotwo{An additional case is simulated to illustrate the applicability to dynamic current conditions. \revtwothree{Conceptually, the training procedure is the same as when using a constant-current condition. The only difference is that the residual of the Li concentrations boundary conditions (Eq.~\ref{eq:current_and_flux}), minimized at the collocation points, is time-dependent.} A discharge C-rate is assumed to follow a sinusoidal profile given by $2 - 2 \operatorname{sin}( 2 \pi t / T)$, where $T$ is the total time interval. Here, $T = 1350$~s given that the average C-rate is equal to 2. The time-varying case is trained with 3000 ADAM SGD epochs and 30000 L-BFGS epochs, and all the other training parameters are held constant.}

\revtwotwo{Two error metrics are computed hereafter: the scaled mean absolute error (Eq.~\ref{eq:errSPM}) and a mean absolute terminal voltage error defined as}
\begin{equation}
    \label{eq:errtv}
    \revtwotwo{\varepsilon_{\rm TV} = \frac{1}{N_{\xi}} \sum_{i \in [1,N_{\xi}]} \left| \phi_{\rm s, c, CC, PINN} - \phi_{\rm s, c, CC, PDE} \right|},
\end{equation}
\revtwotwo{where $\phi_{\rm s, c, CC, PDE}$ is the potential at the cathode current collector obtained from finite difference at the point $i$, $\phi_{\rm s, c, CC, PINN}$ is the predicted potential at the cathode current collector at the point $i$, and $N_{\xi}$ is the number of points over which the error is computed.}

\revtwotwo{Figure~\ref{fig:spm_cc} shows that the effect of the constant current condition on the errors observed is negligible compared to the effect of the architecture or the training regularization procedure. Therefore, it is expected that the analysis in Sec.\ref{sec:spmSurr} would apply to different C-rates other than the 2C discharge.}

\section{\revtwothree{Applicability to sharply varying current conditions}}
\label{sect:SPM_othersharp_Appendix}

\begin{figure}[th!]
	\centering
        \includegraphics[width=3.176in]{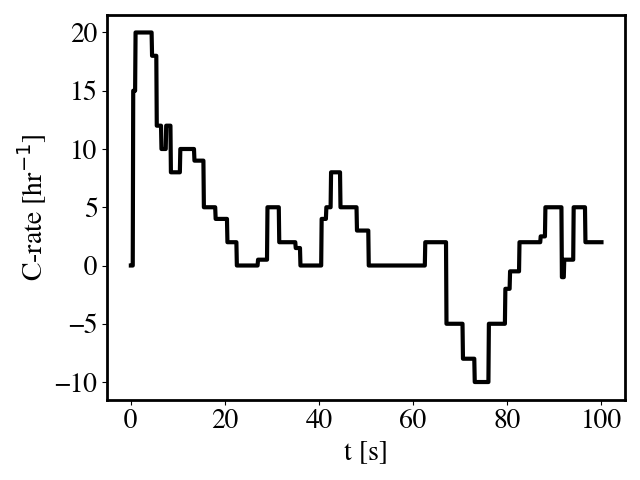}
	\caption{\revtwothree{Hybrid electric vehicle drive cycle considered.}}
	\label{fig:ev_drivecycle}
\end{figure}

\revtwothree{The PINN architecture and training procedure allows for fast predictions over arbitrary horizon times of the battery state (e.g., voltage) response. In Appendix~\ref{sect:SPM_otherCC_Appendix}, the approach is shown to be successful when using a relatively small number of trainable parameters ($9,004$ in all the cases shown in Appendix~\ref{sect:SPM_otherCC_Appendix}). In this section, it is shown that a small number of trainable parameters is appropriate only if the state variables vary smoothly over time and space. To demonstrate this, a hybrid electric vehicle drive-cycle is considered in this section and is shown in Fig.~\ref{fig:ev_drivecycle}\footnote{https://www.comsol.com/model/1d-lithium-ion-battery-drive-cycle-monitoring-19133}. Compared to the sinusoidal cycle used in Appendix~\ref{sect:SPM_otherCC_Appendix}, the current-demand exhibits high-amplitude and high-frequency variations.}

\begin{figure}[h!]
    \centering
    \includegraphics[width=0.8\linewidth]{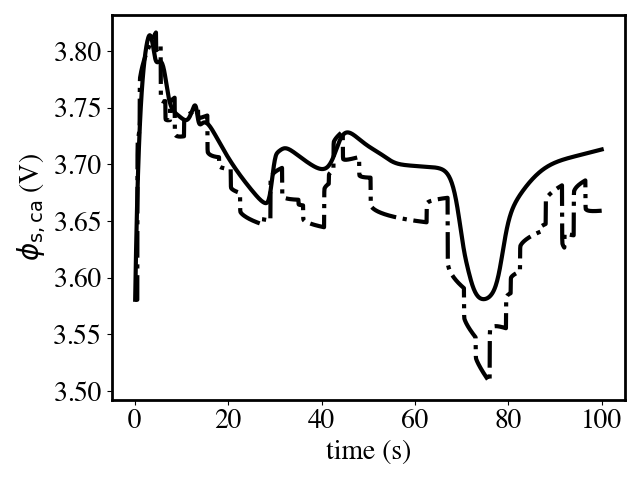}
    \includegraphics[width=0.8\linewidth]{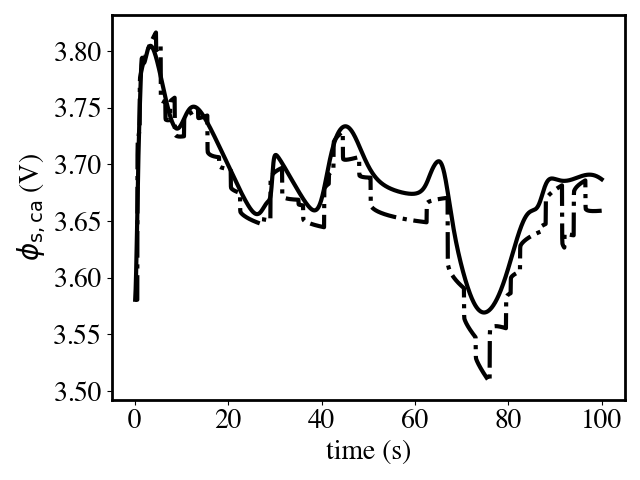}
    \includegraphics[width=0.8\linewidth]{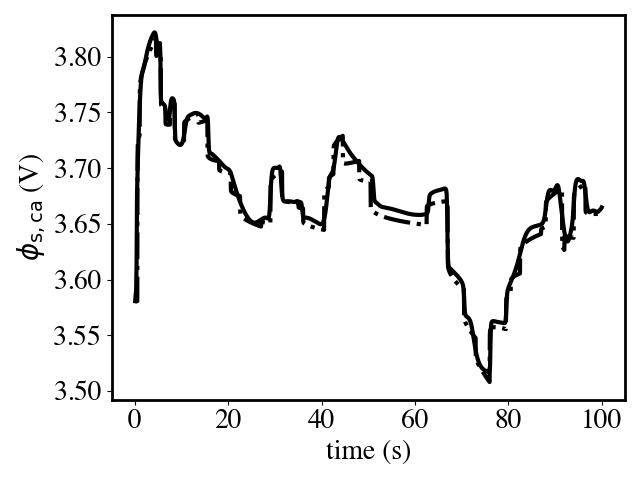}
    \caption{Voltage response predicted by the SPM-PINN surrogates (lines) for a realistic drive cycle for a) \textit{Base}, b) \textit{Col} and c) \textit{Col+Par} as compared to the SPM-PDE solution (dashes).}
    \label{fig:voltResp}
\end{figure}

\revtwothree{To study the developed PINN applicability to high-frequency current demands, three types of PINN architectures are considered.  First an SPM-PINN surrogate (referred to as \textit{Base}) is developed using $9,004$ parameters (same architecture as described in Table~\ref{tab:architecture}) and using $10,240$ collocation points evenly distributed between the interior and the boundary domains. Second, an SPM-PINN surrogate (referred to as \textit{Col}) is trained with 5$\times$ more collocation points ($51,200$ points) and the same number of trainable parameters ($9,004$). Third, a larger SPM-PINN surrogate (referred to as \textit{Col+Par}) is trained ($44,644$ parameters) by doubling the number of neurons per layer (from $20$ to $40$) and doubling the number of gradient pathology blocks (from 2 to 4), and by using the maximum number of collocation points ($51,200$). All models are trained only with a physics loss for $3,000$ SGD epochs and $50,000$ LBFGS epochs. Table~\ref{tab:drivecyc} summarizes the training conditions and average the terminal voltage error $\varepsilon_{\rm TV}$ over the drive-cycle. Figure~\ref{fig:voltResp} illustrates the predicted voltage response from each surrogate as compared to a PDE solution.}

\begin{table}[h!]
\begin{center}
\caption{\revtwothree{Training conditions and results for the hybrid electric vehicle drive cycle.}}
\label{tab:drivecyc}
\begin{tabular}{||c | c | c | c||} 
 \hline
 Name & $\#$ Col. pts & $\#$ Parameters & $\varepsilon_{\rm TV}$ [$\rm{mV}$].\\ [0.5ex] 
 \hline\hline
 Base & $10,240$ & $9,004$ & $34.35$\\ 
 \hline
 Col & $51,200$ & $9,004$ & $20.12$\\
 \hline
 Col+Par & $51,200$  & $44,644$  & $6.08$\\
 \hline
\end{tabular}
\end{center}
\end{table}

\revtwothree{Figure~\ref{fig:voltResp} shows that the smaller PINN models tend to smooth out the cathode current-collector potential $\phi_{s,c}$ temporal gradients, which eventually results in error accumulation. Adding collocation points (compare case \textit{Col} and case \textit{Base}) improves the predictions, but still results in overly smooth predictions. The third case (\textit{Col+Par}, Fig.~\ref{fig:voltResp}c) provides the best predictions thanks to a more expressive network, resulting in terminal voltage errors that are near that of the constant-current results (Appendix~\ref{sect:SPM_otherCC_Appendix}). Therefore, if complex drive cycles are used, the expressiveness of the network needs to be enhanced by increasing the number of trainable parameters. Additionally, as the number of sharp current variations increases, strategies other than increasing the PINN expressiveness may need to be considered.  For example, strategies have been proposed previously that do not encode the entire time-series of the state variables~\cite{li2020fourier,venturi2023svd,hochreiter1997long,zheng2023state}.}

\end{document}